\pdfoutput=1

\documentclass[11pt]{article}

\usepackage{acl}
\usepackage{times}
\usepackage{latexsym}

\usepackage[T1]{fontenc}

\usepackage[utf8]{inputenc}

\usepackage{microtype}

\usepackage{hyperref}
\usepackage{amssymb}
\usepackage{amsmath}
\usepackage{booktabs}
\usepackage{multirow}
\usepackage{xspace}
\usepackage{tablefootnote}
\usepackage{caption}
\usepackage{subcaption}
\usepackage{array}
\usepackage{graphicx}
\usepackage{dsfont}

\definecolor{tw_color}{RGB}{51, 124, 195}
\definecolor{ow_color}{RGB}{195, 4, 4}

\newcommand{\tw}[1]{\textcolor{tw_color}{\textbf{#1}}}
\newcommand{\ow}[1]{\textcolor{ow_color}{\textbf{\underline{#1}}}}

\makeatletter
\DeclareRobustCommand\onedot{\futurelet\@let@token\@onedot}
\def\@onedot{\ifx\@let@token.\else.\null\fi\xspace}
 \def\Eg{\emph{E.g}\onedot}
\def\ie{\emph{i.e}\onedot} \def\Ie{\emph{I.e}\onedot}
 
 \def\vs{\emph{vs}\onedot}

\DeclareRobustCommand\rightbracket{\futurelet\@let@token\@rightbracket}
\def\@rightbracket{\ifx\@let@token]\else]\null\fi\xspace}
\def\mask{{\sc[mask}\rightbracket}
\def\firstpn{dependence measure\xspace}
\def\Firstpn{Dependence measure\xspace}
\def\FirstPN{Dependence Measure\xspace}

\def\secondpn{effectiveness measure\xspace}
\def\Secondpn{Effectiveness measure\xspace}
\def\SecondPN{Effectiveness Measure\xspace}

\def\firstrq{Which association do PLMs depend on to capture factual knowledge?\xspace}
\def\secondrq{Is the association on which PLMs depend effective in capturing factual knowledge?\xspace}

\newtheorem{defn}{Definition}
\newtheorem{observation}{Observation}
\newtheorem{subquestion}{Research Question}

\title{How Pre-trained Language Models Capture Factual Knowledge? A Causal-Inspired Analysis}

\author{
Shaobo Li$^{1}$\thanks{\hspace{0.3pt} Authors contribute equally.},
Xiaoguang Li$^{2 \hspace{0.3pt} * \hspace{0.3pt}}\thanks{\hspace{0.5pt} Corresponding authors: sunchengjie@hit.edu.cn, lixiaoguang11@huawei.com}$,
Lifeng Shang$^{2}$,
Zhenhua Dong$^{2}$,\\
\textbf{Chengjie Sun$^{1 \dag}$, Bingquan Liu$^{1}$, Zhenzhou Ji$^{1}$, Xin Jiang$^{2}$ and  Qun Liu$^{2}$}\\
$^{1}$Harbin Institute of Technology\\
$^{2}$Huawei Noah's Ark Lab\\
{shli@insun.hit.edu.cn},
{\{sunchengjie, liubq, jizhenzhou\}@hit.edu.cn}\\
{\{lixiaoguang11, shang.lifeng, dongzhenhua, Jiang.Xin, qun.liu\}@huawei.com}\\
}

\begin{document}
\maketitle
\begin{abstract}

Recently, there has been a trend to investigate the factual knowledge captured by Pre-trained Language Models (PLMs). Many works show the PLMs' ability to fill in the missing factual words in cloze-style prompts such as ``Dante was born in \mask.'' However, it is still a mystery how PLMs generate the results correctly: relying on effective clues or shortcut patterns? We try to answer this question by a causal-inspired analysis that quantitatively measures and evaluates the word-level patterns that PLMs depend on to generate the missing words. We check the words that have three typical associations with the missing words: \textit{knowledge-dependent}, \textit{positionally close}, and \textit{highly co-occurred}. Our analysis shows: (1) PLMs generate the missing factual words more by the positionally close and highly co-occurred words than the knowledge-dependent words; (2) the dependence on the knowledge-dependent words is more effective than the positionally close and highly co-occurred words. Accordingly, we conclude that the PLMs capture the factual knowledge ineffectively because of depending on the inadequate associations.

\end{abstract}

\section{Introduction}

Do Pre-trained Language Models~(PLMs) capture factual knowledge? LAMA benchmark~\cite{lama} answers this question by quantitatively measuring the factual knowledge captured in PLMs: query PLMs with cloze-style prompts such as ``Dante was born in \mask?'' Filling in the mask with the correct word ``Florence'' is considered a successful capture of the corresponding factual knowledge. The percentage of correct fillings over all the prompts can be used to estimate the amount of factual knowledge captured. PLMs show a surprisingly strong ability to capture factual knowledge in such probings \cite{JiangXAN20,shin2020autoprompt,optim-prompt}, which
elicits further research on a more in-depth question \cite{lanka,knowledge-consistency}: How do PLMs capture the factual knowledge?  In this paper, we try to answer this question with a two-fold analysis:

\begin{subquestion}
     \firstrq
     \label{rq:1}
\end{subquestion}
\begin{subquestion}
    \secondrq
    \label{rq:2}
\end{subquestion}

\begin{figure}[t]

\centering
\small
\begin{tabular}{p{0.9\columnwidth}}
\toprule
\textbf{Knowledge-Dependent:} \\
\tw{Columbus} born between 25 August and 31 October 1451, \tw{died} \ow{20 May 1506} was an Italian explorer. \\ \midrule
\textbf{Positionally Close:} \\
Columbus born between 25 August and 31 October 1451, \tw{died} \ow{20 May 1506} \tw{was} an Italian explorer. \\ \midrule
\textbf{Highly Co-occurred:} \\
\tw{Columbus} born between 25 August and 31 October 1451, died \ow{20 May 1506} was an Italian \tw{explorer}. \\
\bottomrule
\end{tabular}
    \caption{The associations we investigated. The underlined words are the missing words that need to be generated. The bold words, which hold specific associations with the missing words, are considered as the word-level patterns that PLMs may use to generate the missing words.}
\label{fig:association}

\end{figure}
\begin{figure*}[thbp]

\centering
\begin{subfigure}[b]{0.44\textwidth}
    \centering
    \includegraphics[width=\textwidth]{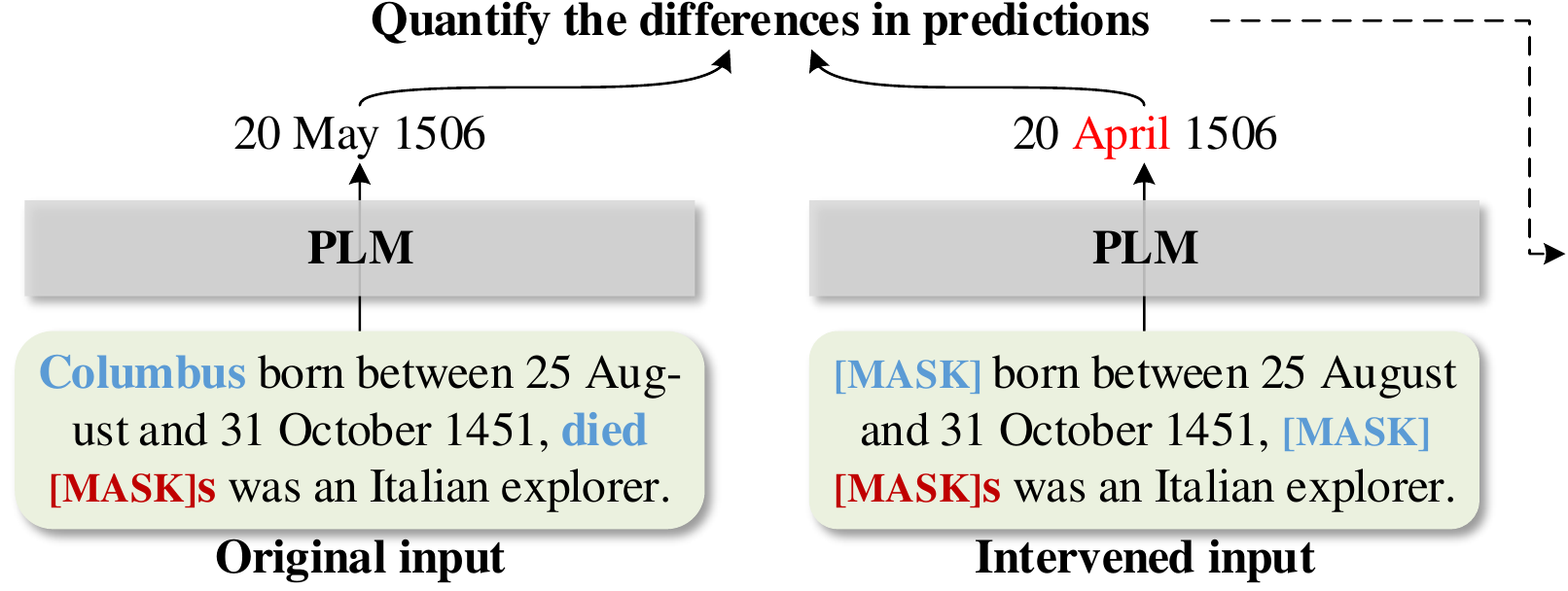}     
    \caption{\Firstpn.}
    \label{fig:overview-1}
\end{subfigure}
\ 
\begin{subfigure}[b]{0.48\textwidth}
    \centering
    \includegraphics[width=\textwidth]{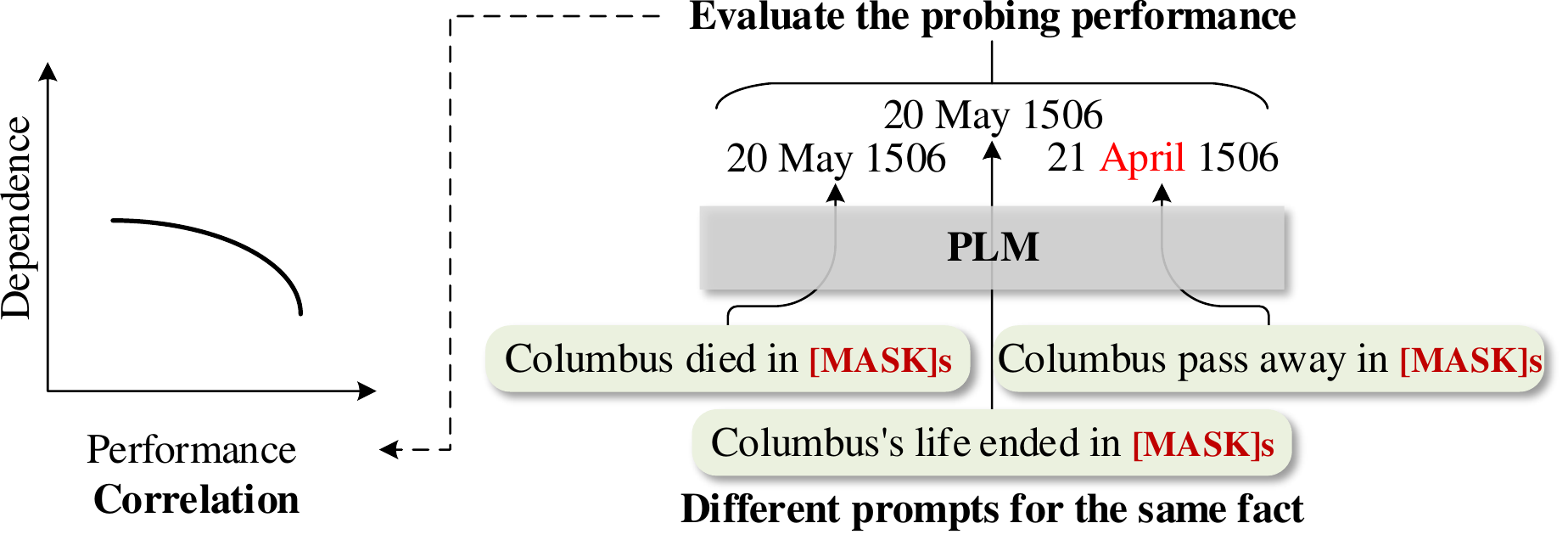}     
    \caption{\Secondpn.}
    \label{fig:overview-2}
\end{subfigure}

\caption{The overview of the proposed analysis framework. The \firstpn quantifies how much the PLMs depend on each association to capture factual knowledge when per-training. The \secondpn evaluates whether the dependence on an association is good for the factual knowledge performance in probing.}
\label{fig:overview}
\end{figure*}
\noindent We use \textit{association} to refer to the explicit association between the missing words and the remaining words in the context. We define three typical associations between words. Figure~\ref{fig:association} illustrates these associations in a mask-filling sample.
\begin{defn}
\textbf{Knowledge-Dependent (KD)}: According to a Knowledge Base~(KB), the missing words can be deterministically predicted when providing the remaining words.
\end{defn}
\begin{defn}
\textbf{Positionally Close (PC)}: The remaining words are positionally close to the missing words.
\end{defn}
\begin{defn}
\textbf{Highly Co-occurred (HC)}: The remaining words have a higher co-occurrence frequency with the missing words.
\end{defn}
Question~\ref{rq:1} investigates how much PLMs depend on a specific group of remaining words to predict the missing words in pre-training samples. We select the remaining words to be investigated according to their association with the missing words. We propose a causal-inspired method to quantify the word-level dependence in each sample. The average dependence on the remaining words that hold the same association with the missing words over all the samples indicates how PLMs rely on this association to predict the missing words. We refer to this average dependence as dependence on the association. The above analysis is named \firstpn.

In Question \ref{rq:2}, we reveal the effectiveness of dependence by the correlation between the quantified dependence on associations and the factual knowledge capturing performance. The performance is probed with additionally crafted cloze-style prompts\cite{knowledge-consistency}. The more the dependence on an association positive correlates with the probing performance, the more effective this association is. We refer to the second analysis as effectiveness measure.
According the experiment results, we have the following observations:

\begin{observation}
The PLMs depend more on the positional close and highly co-occurred associations than the knowledge-dependent association to capture factual knowledge.
\end{observation}
\begin{observation}
Depending on the knowledge-dependent association is more effective for factual knowledge capture than positional close and highly co-occurred associations.
\end{observation}
By connecting the two observations, we can answer the question of ``\textit{how PLMs capture factual knowledge}'' as: \textbf{The PLMs are capturing factual knowledge ineffectively since the PLMs depend more on the PC and HC association than the KD association.}

The contribution of this paper can be summarized as follows: (1) We quantify the word-level dependence for mask filling with a causal-inspired method, revealing the word-level patterns that PLMs use to predict the missing words quantitatively. (2) We compare the effectiveness of the dependence on different associations, which provides direct insights for improving PLMs for factual knowledge capture. (3) This paper introduces causal theories into PLMs by formulating the effect measurement process in mask language modeling. It paves the path to measure the causal effects between entities or events described in natural language.

\begin{table*}[htbp] 
\centering
\small
\begin{tabular}{ 
>{\centering}m{0.12\textwidth}
>{\centering}m{0.25\textwidth}
>{\centering}m{0.25\textwidth}
>{\centering\arraybackslash}m{0.25\textwidth}}
\toprule
\textbf{Association} &
\multicolumn{1}{c}{\textbf{Knowledge-Dependent}} &
\multicolumn{1}{c}{\textbf{Positionally Close}} &
\multicolumn{1}{c}{\textbf{Highly Co-occurred}} \\\midrule
\textbf{Input}     & 
    \tw{$W_t$} born between 25 August and 31 October 1451, \tw{$W_t$} \ow{$W_o$} was an Italian explorer.  &
    Columbus born between 25 August and 31 October 1451, \tw{$W_t$} \ow{$W_o$} \tw{$W_t$} an Italian explorer.  &
    \tw{$W_t$} born between 25 August and 31 October 1451, died \ow{$W_o$} was an Italian \tw{$W_t$}.  \\\midrule
\textbf{SCM} & 
{\includegraphics[height=55pt]{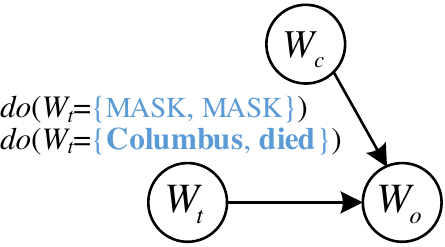}} &
{\includegraphics[height=55pt]{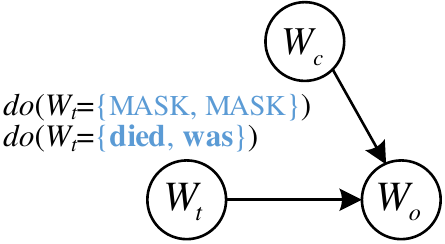}} &
{\includegraphics[height=55pt]{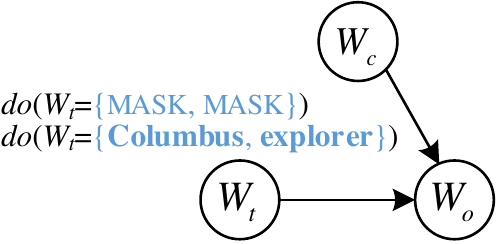}} \\

\bottomrule
\end{tabular}

\caption{To analyze the dependence of associations, we do interventions on treatment words to reveal their causal effects on the outcome words.}
\label{tab:DAG}
\end{table*}

\section{Method}

\subsection{Overview}
\label{sec:overview}
We take a quick overview of our two-fold analysis with a running example in Figure~\ref{fig:overview}.
Figure~\ref{fig:overview-1} illustrates how to measure the dependence on the remaining words ``Columbus'' and ``died'' when predicting the missing words ``20 May 1506.'' We let the PLM generate the missing words based on the original input first, then mask the remaining words in the input and let PLMs generate again. The difference between these two predictions is quantified and used to measure the dependence. The remaining and missing words hold the knowledge-dependent association in this sample. We repeat this measure on all the samples whose remaining and missing words have the KD association. Then the dependence on the KD association can be estimated by the average of the quantified difference. 

Figure~\ref{fig:overview-2} measures the effectiveness of the dependence on each association by calculating the correlation coefficient between the dependence and the probing performance. Following \cite{lama, knowledge-consistency}, the probing performance is indicated by the prediction accuracy and consistency when querying on the same fact with different prompts. Since the dependence on associations are quantified in the \firstpn, we can calculate the correlation coefficient between the dependence and performance over all the samples. Their correlation measures whether the dependence on an association is harmful or beneficial to the performance, showing the effectiveness of the dependence on an association quantitatively.

\noindent\textbf{Section Outline}
We organize the rest of this section as follows. In Section~\ref{sec:estimate_SCM}, we formalize how we quantify the dependence with the causal effect estimation. Section~\ref{sec:ass_des} gives detail about how we build the probing samples for different associations. Section~\ref{sec:performace_metric} introduces the metrics we used to indicate the performance of factual knowledge capture. Section~\ref{sec:correlate} describes the details about the \secondpn of associations.

\subsection{Quantify the Dependence on Associations}
\label{sec:quantify_dep}

\subsubsection{Causal Effect Estimation for PLMs}
\label{sec:estimate_SCM}

To study the causal effect of the different input words, we build a Structured Causal Model (SCM) for the missing words generation process and apply interventions on some input words to estimate their effect quantitatively. We consider the missing words as \textbf{outcome words} and the remaining words that hold a certain association~(e.g. positionally close) with the outcome words as \textbf{treatment words}. Then, we can formally represent the word generation process with SCM, as the following structural equations:

\begin{equation}
\begin{split}
    &w_c=f(\mathbb{I}), w_t={\rm PLM}(w_c)
    \\
    &w_o={\rm PLM}(w_c,w_t).
\end{split}
\label{eq:structral-equations}
\end{equation}

Separate the words in a sentence into three groups: treatment words $W_t$, outcome words $W_o$, and context words $W_c$ (specified by $w_t$, $w_o$, and $w_c$ respectively). Equation \ref{eq:structral-equations} formulates the following data generation process: (1)~Sample a sentence from the natural text space $\mathbb{I}$ and get the context words $w_c$ using function $f$. (2)~Generate the treatment words $w_t$ by the PLM based on $w_c$ only. (3)~Generate outcome words $w_o$ based on both the $w_c$ and $w_t$.

To obtain the quantitative causal effect of treatment words $W_t$ on outcome words $W_o$, we apply the do-calculus $do()$~\cite{pearl2009causality} on treatment words $W_t$ to introduce interventions for estimating the causal effect. $do()$ denotes the operation of forcibly setting the value of $W_t$. Then the causal effect of $W_t$ on $W_o$ can be estimated by the Average Treatment Effect~(ATE) \cite{rubin1974estimating}: 
\begin{equation}
\begin{split}
    &\mathbb{E}\left[ P(W_o|do(W_t=\hat{w}_t)) \right] \\
    -&\mathbb{E}\left[ P(W_o|do(W_t=w_m)) \right].
\end{split}
\end{equation}
Accordingly, we define ATE for PLMs as:
\begin{equation}
\begin{split}
    \tau &=\sum_{\mathbb{I}}{\rm PLM}\left(do(W_t=\hat{w}_t),w_c\right)P(s) \\
    &-\sum_{\mathbb{I}}{\rm PLM}\left(do(W_t=w_m),w_c\right)P(s),
\end{split}
\label{eq:ate}
\end{equation}
where $\hat{w}_t$ is the ground truth of the treatment words $W_t$ (the original value of $W_t$ without intervention). $w_m$ is the intervention value~(several \mask{}s) for $W_t$, and we use it to simulate removing the ground-truth value $\hat{w}_t$ from the input. $P(s)$ denotes the probability of selecting the sample $s$ that consists of $w_t$, $w_o$, and $w_c$ from $\mathbb{I}$. $\rm PLM(\cdot,\cdot)$ denotes the output of PLMs with certain input. Table~\ref{tab:DAG} illustrates the interventions on the SCM for different associations.

The raw output of PLMs is a probability distribution over fixed vocabulary. We transform the output into reciprocal rank to quantify the differences:

\begin{equation}
{\rm PLM}_{k}{(w_t,w_c)} =  
\begin{cases}
     \frac{1}{rank_{\hat{w}_o}}, & \text{if } rank_{\hat{w}_o} \leq k\\
     0, & \text{otherwise}
\end{cases}.
\label{eq:mrr-ate}
\end{equation}
$\hat{w}_o$ is the ground-truth outcome words. $rank_{\hat{w}_o}$ is the rank position of $\hat{w}_o$ according to the generation probability of $\hat{w}_o$ output by ${\rm PLM} (w_c, w_t)$. We set $k$ to 100 and use ${\rm PLM_{100}}$ to replace ${\rm PLM}$ in Equation~\ref{eq:ate} to calculate ATE. The ATE reflects the effect of $W_t$ on the prediction of $W_o$, it can be regarded as a quantitative estimation of how much PLMs depend on $W_t$ when generating $W_o$. 

\subsubsection{Mark words by Associations}
\label{sec:ass_des}

Wikipedia is a rich source of knowledge \cite{wikipedia-knowledge-source-1,wikipedia-knowledge-source-2}, and most of the PLMs nowadays have been pre-trained on Wikipedia \cite{BERT,roberta,ALBERT}, so we take Wikipedia sentences as pre-training samples to construct the probing samples for \firstpn. We probe the mask-filling on these sentences to analyze what PLMs based on when capturing factual knowledge in pre-training.

The outcome we want to observe is the predictions of factual words in the sentences. In order to locate the factual words, we align each sentence with a triplet (\textit{subject}, \textit{predicate}, \textit{object}) in the KB. The words that correspond to the \textit{object} serve as outcome words $W_o$  for observation, and the remaining words that hold an explicit association with $W_o$ are marked as treatment words $W_t$  for intervention. For different associations, the $W_t$ is identified as:
\begin{enumerate}
    \item \textit{Knowledge-Dependent}: all the remaining words correspond to the \textit{predicate} and \textit{object} in the same triplet with the $W_t$.
    \item \textit{Positionally Close}: the remaining words closest to $W_o$. 
    \item \textit{Highly Co-occurred}: the remaining words that have higher Pointwise Mutual Information~(PMI)~\cite{PMI} with $W_o$. The PMI is calculated over all the Wikipedia sentences using the following equation:
    \begin{equation}
    {\rm PMI}(w_i;\hat{w}_o) = \frac{P(w_i|\hat{w}_o)}{P(w_i)},
    \end{equation}
    where $\hat{w}_o$ is a group of words (a span) and $w_i$ is a single word.
    \item We further define a \textit{Random}~(\textbf{R}) association, where the $W_t$ are randomly selected remaining words. It provides some empirical support for how much the modifications in the context affect the mask-filling output. 
\end{enumerate}

Accordingly, one sentence yields four probing samples for the four associations, respectively. The four probing samples share the same $W_o$ but use different words as $W_t$ to show the dependence on different associations when predicting the same $W_o$. We preserve that the number of words in $W_t$ is the same among different associations. For example, if there are two words used as $W_t$ by the KD association, we select the top two closest words with $W_o$ as $W_t$ for PC, and the words have the top two PMI with $W_o$ for HC. We can obtain a set of probing samples for each association. The sample sets for different associations source from the same set of sentences and have the same size.

\subsection{Measure the Effectiveness of Associations}
\label{sec:eff_chk}
This section investigates which association can lead to better performance on factual knowledge capture. We first define the metrics to evaluate the performance, then we measure the effectiveness of an association by relating the dependence on this association with probing performance. 

\subsubsection{Metrics for Factual Knowledge Probing}
\label{sec:performace_metric}
Section \ref{sec:quantify_dep} uses the original Wikipedia sentence as pre-training samples to quantify the dependence PLMs used to capture the corresponding fact in pre-training. The performance of capturing the corresponding fact is probed by having PLMs fill masks on crafted quires. We construct these queries by instantiating templates on triplets \cite{lama}. $T_i(s)$ denotes the $i$-th query for the fact corresponds to $s$. The accuracy ${\rm mrr}$ of capturing this fact is obtained by averaging over all the predictions obtained with different queries:
\begin{equation}
    {\rm mrr}\left(s\right) = \frac{1}{n}\sum_{i}^{n}{\rm PLM}_{k}({T_i(s)}),
    \label{eq:correct_metric}
\end{equation} 
${\rm PLM}_{k}({T_i(s)})$ denotes the reciprocal rank of the ground truth in the PLM's output for query $T_i(s)$, it is defined in Equation~\ref{eq:mrr-ate}.

The consistency of the capture is indicated by the percentage of the pairs of queries that have the same result \cite{knowledge-consistency}:
\begin{equation}
    {\rm con}\left(s\right) = \frac{\sum_{i\ne j} \mathds{1}_{
        {\rm PLM}({T_i(s)})={\rm PLM}(T_j(s))}
    } {n(n-1)}
    .
    \label{eq:consist_metric}
\end{equation}
There are $n$ different queries on every fact, and we can get $\binom{n}{2}=n(n-1)$ pairs of predictions in total. ${\rm PLM}({T_i(s)})$ denotes the top-1 output for the query $T_i(s)$. The value of $\mathds{1}_{{\rm PLM}({T_i(s)})={\rm PLM}(T_j(s))}$ is an indicator function that takes the value 1 if the PLMs returns identical at top-1 for ${T_i(s)}$ and ${T_j(s)}$ and 0 otherwise. The PLMs are better on the consistency metric if they keep the predictions consistent when queries only vary on the surface forms. \Eg, the two queries ``Dante was born in \mask'' and ``The birthday of Dante is \mask'' should return the same results. 

Finally, we evaluate the factual knowledge capture performance by jointly examining the accuracy and consistency \cite{knowledge-consistency}:
\begin{equation}
    {\rm test}(s) = {\rm mrr}\left(s\right)\cdot{\rm con}\left(s\right)
    \label{eq:joint_metric}
\end{equation}
${\rm test}(s)$ measures the probing performance on template-based queries. We also define a metric to measure how well the PLMs memorize the missing words in pre-training samples (Wikipedia sentences):
\begin{equation}
    {\rm train}(s) = {\rm PLM}_k(s).
    \label{eq:correct_metric_train}
\end{equation}

\subsubsection{Correlate Performance with Dependence}
\label{sec:correlate}
We have quantified the dependence on each association and defined the metrics for probing performance in the above sections. We then calculate the Pearson correlation coefficient~\cite{Pearson-coefficient} between dependence and probing performance to reveal the effectiveness of different associations. An association is considered more effective if the probing performance positively correlates with its dependence more.

Because only part of the facts has available templates, the samples in the \firstpn without templates are ignored in the calculation. The factual knowledge captured by different PLMs may vary significantly due to the differences in model scale, pre-training data, or other settings. To make the correlation coefficient comparable between different PLMs, we calculate the correlation only on the factual knowledge gathered correctly by the PLM. \Ie, only the pre-training samples with ${\rm train}(s)={\rm PLM}_k(s)=1$ are involved.

\begin{table}[bt]
\small
\centering
\begin{tabular}{lc}
\toprule
\multicolumn{2}{c}{\textbf{Sample in \FirstPN}}   \\
\midrule
\# Wikipedia sentences                           & 4,779,753   \\
\# Different triplets                  & 3,795,229   \\
\# Different predicates                & 565         \\
\# Sentences with synthetic templates            & 1,119,875   \\

\midrule
\multicolumn{2}{c}{\textbf{Queries in \SecondPN}}  \\
\midrule
\# Template-based queries                         & 7,645,635    \\
\# Different triplets              & 654,112      \\
\# Different predicates            & 38           \\
\# Different templates             & 328          \\


\bottomrule
\end{tabular}
\caption{Statistics of the probing data.}
\label{tab:dataset_stat}
\end{table}

\section{Experiments and Discussions}
\begin{table*}[tbp]

    \small
    \centering
    \begin{tabular}{lccccc}
    \toprule
    \multirow{2}{*}{\textbf{Model}} & \multirow{2}{*}{\textbf{Accuracy}} & \multicolumn{4}{c}{\textbf{Dependences on Associations} ($k=100$)}                                 \\ \cmidrule{3-6} 
                                    & &  \textbf{KD}    &  \textbf{PC} &  \textbf{HC} &  \textbf{R} \\
    \midrule
    
    BERT\textit{-base-cased}        &   0.3623 &            0.1585 &             0.4085 &             0.1779 &  0.1081 \\
    BERT\textit{-large-cased}       &   0.3692 &            0.1603 &             0.4113 &             0.1791 &  0.0996 \\
    BERT\textit{-large-cased-wwm}   &   0.5030 &            0.1384 &             0.4477 &             0.2305 &  0.1072 \\
    SpanBERT\textit{-large}         &   0.5223 &            0.1351 &             0.3679 &             0.2383 &  0.1157 \\
    RoBERTa\textit{-base}           &   0.3511 &            0.1352 &             0.3926 &             0.2093 &  0.1053 \\
    RoBERTa\textit{-large}          &   0.4276 &            0.1360 &             0.3962 &             0.2162 &  0.0985 \\
    \midrule
    BERT\textit{-large-uncased-wwm} &   0.5035 &            0.1410 &             0.4350 &             0.2290 &  0.1089 \\
    
    ALBERT\textit{-xxlarge-v2}      &   0.4758 &            0.2852 &             0.4338 &             0.3801 &  0.2704 \\
    

    \bottomrule
    \end{tabular}
    \caption{The quantified dependence on associations. \textbf{Accuracy} denotes the performance of filling in the masks in pre-training samples. The PLMs use cased and uncased vocabularies are separated.}
    \label{tab:ate}
\end{table*}

\begin{figure*}[th]
     \centering
     \begin{subfigure}[b]{0.48\columnwidth}
         \centering
         \includegraphics[width=\textwidth]{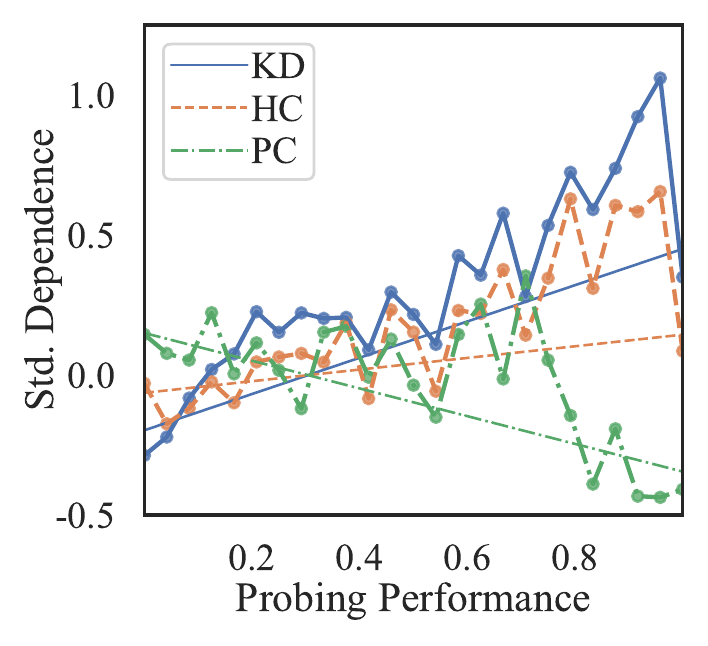}
         \caption{BERT\textit{-large-cased-wwm}}
         \label{fig:ate_and_query_bert_large_uncased_wwm}
     \end{subfigure}\quad
     \begin{subfigure}[b]{0.48\columnwidth}
         \centering
         \includegraphics[width=\textwidth]{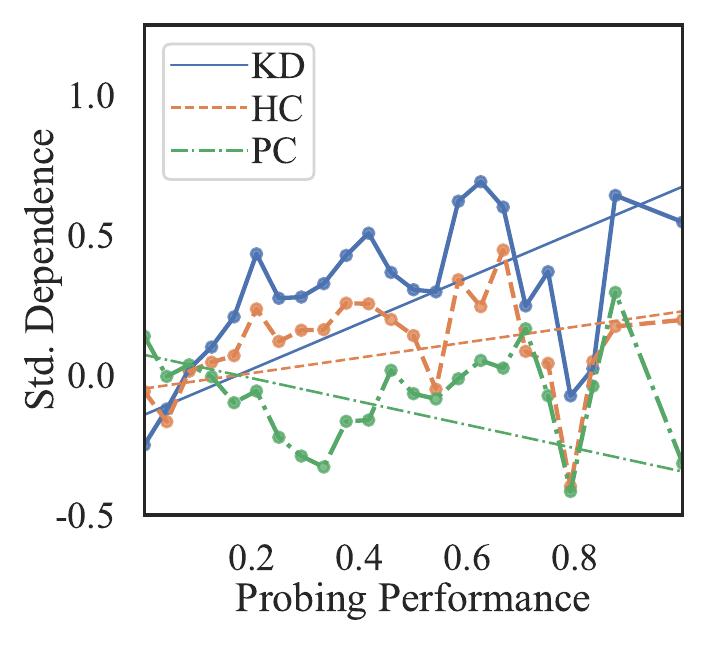}
         \caption{RoBERTa\textit{-large}}
         \label{fig:ate_and_query_roberta_large}
     \end{subfigure}\quad
     \begin{subfigure}[b]{0.48\columnwidth}
         \centering
         \includegraphics[width=\textwidth]{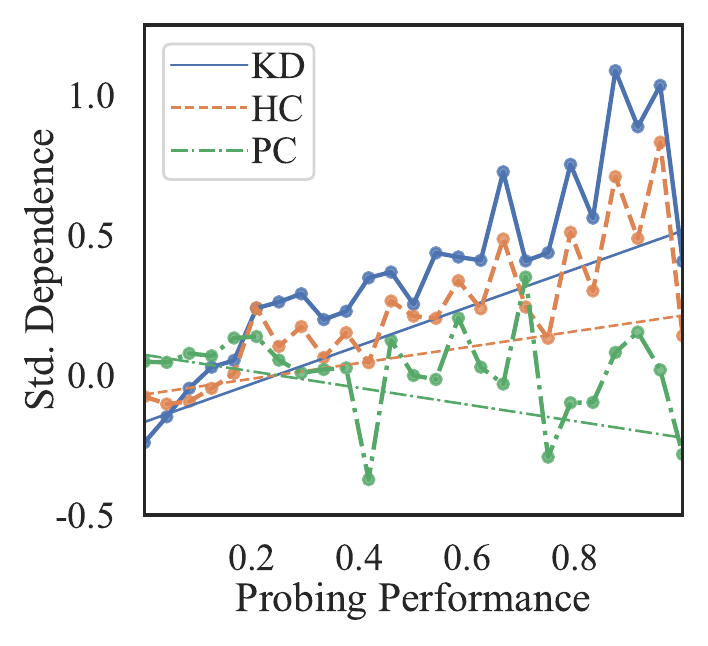}
         \caption{SpanBERT\textit{-large}}
         \label{fig:ate_and_query_spanbert_large}
     \end{subfigure}\quad
     \begin{subfigure}[b]{0.48\columnwidth}
         \centering
         \includegraphics[width=\textwidth]{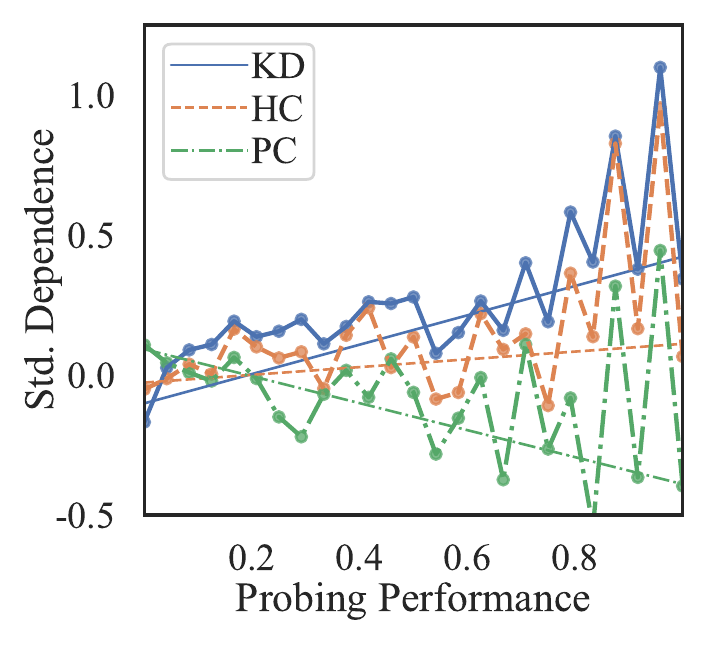}
         \caption{ALBERT\textit{-xxlarge-v2}}
         \label{fig:ate_and_query_albert_xxlarge_v2}
     \end{subfigure}

\caption{The correlations between the dependence on associations and the probing performance on factual knowledge capture.}
\label{fig:ate_and_query}
\end{figure*}

\begin{table*}[thb]
\centering
\tiny
\begin{tabular}{cp{0.8\linewidth}cc}
    \toprule

    &\textbf{Pre-training samples for the \firstpn (Wikipedia Sentence)} & \textbf{Dep.} \\\cmidrule{2-3}
    &\textbf{KD:} Kimwenza is a community in the Democratic Republic of the \tw{Congo} in the Mont Ngafula commune in the south of the \tw{capital}, \ow{Kinshasa} .                   & \textbf{0.8564} \\

    \multirow{7}{*}{\textbf{Case 1}}&\textbf{PC:} Kimwenza is a community in the Democratic Republic of the Congo in the Mont Ngafula commune in the south of the capital\tw{,} \ow{Kinshasa} \tw{.}                   & 0.0000 \\
    &\textbf{HC:} \tw{Kimwenza} is a community in the Democratic Republic of the Congo in the Mont \tw{Ngafula} commune in the south of the capital, \ow{Kinshasa} . & 0.0000 \\
    
    \cmidrule{2-3}
    & \textbf{Template-based Queries for the \secondpn}            & \textbf{MRR} \\\cmidrule{2-3}
    & The capital of Congo is \ow{Kinshasa} .  &  1.0\\
    & \ow{Kinshasa} is the capital of Congo .  &  1.0\\
    & Congo's capital is \ow{Kinshasa} .       &  1.0\\
    
    \midrule

    &\textbf{Pre-training samples for the \firstpn (Wikipedia Sentence)} & \textbf{ATE} \\\cmidrule{2-3}
    &\textbf{KD:} \tw{Drayton} is a hamlet in England, in the county of Northamptonshire, $\ldots$, hundred of Fawsley, ¾ of a mile on the low-lying north western side of the \tw{town} of \ow{Daventry} . & 0.0000 \\
    \multirow{7}{*}{\textbf{Case 2}}& \textbf{PC:} Drayton is a hamlet in England, $\ldots$, in the parish and union of Daventry, hundred of Fawsley, ¾ of a mile on the low-lying north western side of the town \tw{of} \ow{Daventry} \tw{.} & \textbf{0.9496} \\
    & \textbf{HC:} Drayton is a hamlet in England, $\ldots$, in the parish and union of \tw{Daventry}, hundred of \tw{Fawsley}, ¾ of a mile on the low-lying north western side of the \tw{town} of \ow{Daventry} . & 0.8452 \\
    \cmidrule{2-3}

    & \textbf{Template-based Queries for the \secondpn}                          & \textbf{MRR} \\\cmidrule{2-3}
    & Drayton is located in \ow{Daventry} .   & 0.0 \\
    & Drayton is in \ow{Daventry} .           & 0.0 \\
    & Drayton can be found in \ow{Daventry} . & 0.0 \\

\bottomrule
\end{tabular}
\vspace{-7pt}
\caption{Two cases from SpanBERT-\textit{large}. The quantified dependence on associations (denoted by Dep.) and the performance of factual knowledge capture (denoted by MRR).}
\label{tab:cases}
\end{table*}

\subsection{Probing Data and PLMs}
We use the TREX dataset~\cite{TREX}, which aligns KB triplets with Wikipedia sentences, to construct the samples for the \firstpn following the definition in Section~\ref{sec:ass_des}. We employ the templates from \cite{knowledge-consistency} to construct the queries to probe the factual knowledge for the \secondpn. Table \ref{tab:dataset_stat} shows the statistics for the data in the \firstpn and the \secondpn. The PLMs we analyzed include BERT~\cite{BERT}, RoBERTa~\cite{roberta}, SpanBERT~\cite{spanbert}, and ALBERT~\cite{ALBERT}.

\subsection{Dependence on Associations}
The dependence on an association is the average ATE over the probing samples whose treatment words hold that association with the outcome words. Table~\ref{tab:ate} shows the quantified dependence of different associations. The accuracy in Table~\ref{tab:ate} represents the accuracy of recovering the masked factual words in pre-training samples, revealing how well does PLMs memorize the pre-training samples. It is calculated by Equation~\ref{eq:correct_metric_train} with $k=1$.

We find a general trend over all the picked PLMs: the Positionally Close (PC) association takes the dominant effect on the prediction results, the Highly Co-occurred (HC) association comes second, and the least for the Knowledge-Dependent (KD) association. The trend does not change much as increasing the model scale (\textit{large} \vs \textit{base}), using additional training data (RoBERTa \vs BERT), or improving the masking strategy (SpanBERT \vs BERT). Consequently, the accuracy drops the most when perturbing the positionally close words but least on knowledge-dependent words\footnotemark.
\footnotetext{Table~\ref{tab:acc_desc} in the Appendix shows the accuracy decrease when perturbing the different associations.}

The results provide quantitative evidence for Question~\ref{rq:1} of ``\textit{\firstrq}:'' \textbf{PLMs prefer the associations founded with positionally close or the highly co-occurred words to the knowledge-based clues.} It is different from how a conventional KB works, e.g., an object can be retrieved by the corresponding subject and predicate. 

\subsection{Correlations between Dependence and Performance}
\label{sec:correlation}

We show the correlation between association's dependence and the probing performance in Figure~\ref{fig:ate_and_query}. Each point in the figure represents a piece of factual knowledge $s$. We refer to it as a fact for convenience. The horizontal axis indicates ${\rm test}(s)$ for the fact, showing the probing performance of the fact with \secondpn. The vertical axis shows the dependence of associations when capturing this fact, which is quantified by the causal effect estimation defined in Section~\ref{sec:estimate_SCM}.  The straight lines are the regression lines and different associations are shown in different line styles\footnotemark.
\footnotetext{We standardize the quantified value of dependence (denoted as Std. Dependence) and plot a bucket of facts as a single point to show the trends clearly. The correlations without standardization for more PLMs are in Table~\ref{tab:ate_and_query_full}}

As we can see from the results, the dependence on the KD association positively correlates with the probing performance. The dependence on the HC association has a slightly positive correlation or almost has no correlations sometimes (such as ALBERT in Figure~\ref{fig:ate_and_query_albert_xxlarge_v2}). The PC association holds a negative correlation with the performance.

These results can give an empirical answer to ``\textit{\secondrq}:'' \textbf{the more PLMs depend on the Knowledge-Dependent (KD) association, the better PLMs can capture the corresponding factual knowledge.} Meanwhile, \textbf{relying much on the positionally close association is harmful to the probing performance.}

The \firstpn results reveal that the PLMs depend most on the positionally close but least on the knowledge-dependent association. However, in \secondpn, we find that the positionally close association is the most ineffective for factual knowledge capture while the knowledge-dependent association is the most effective. By connecting the two results, we can conclude the answer to the question in the title: \textbf{The PLMs do not capture factual knowledge ideally, since they depend more on the ineffective associations than the effective one}.

\subsection{Case Study}
To illustrate the analysis result intuitively, we show two cases with SpanBERT\textit{-large} in Table~\ref{tab:cases}. The MRR shows the probing performance on the template-based query (calculated by Equation~\ref{eq:correct_metric}). In Case 1, the knowledge-dependent association gains the biggest effect, and the predictions are robust in all the template-based probing. However, the positionally close association takes the main effect in Case 2, while the PLM fails to recall the  word ``England'' with the template-based queries.

\begin{figure*}[th]
     \centering
     \begin{subfigure}[b]{0.48\columnwidth}
         \centering
         \includegraphics[width=\textwidth]{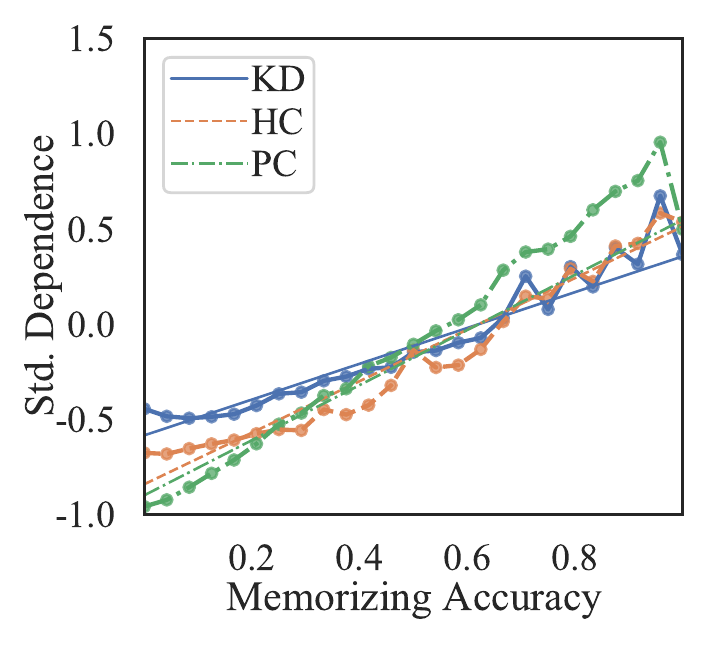}
         \caption{BERT\textit{-large-cased-wwm}}
         \label{fig:ate_and_pt_bert_large_uncased_wwm}
     \end{subfigure}\quad
     \begin{subfigure}[b]{0.48\columnwidth}
         \centering
         \includegraphics[width=\textwidth]{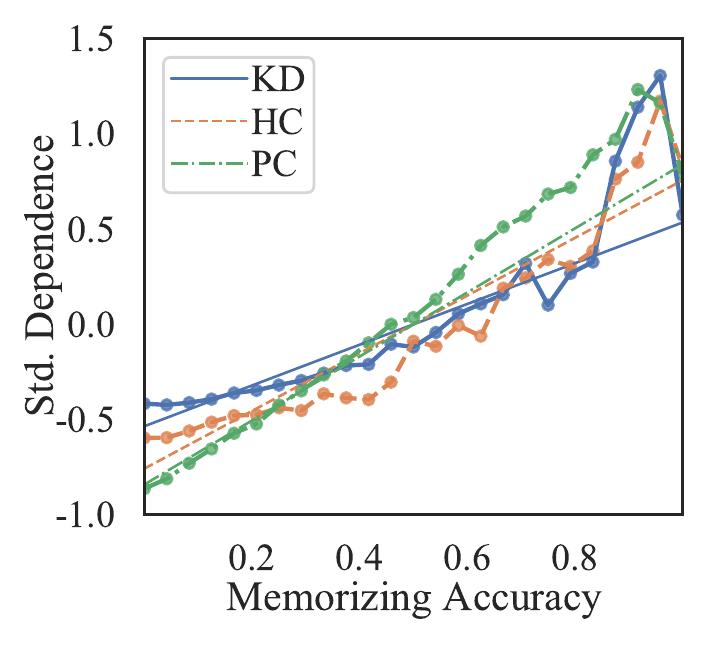}
         \caption{RoBERTa\textit{-large}}
         \label{fig:ate_and_pt_roberta_large}
     \end{subfigure}\quad
     \begin{subfigure}[b]{0.48\columnwidth}
         \centering
         \includegraphics[width=\textwidth]{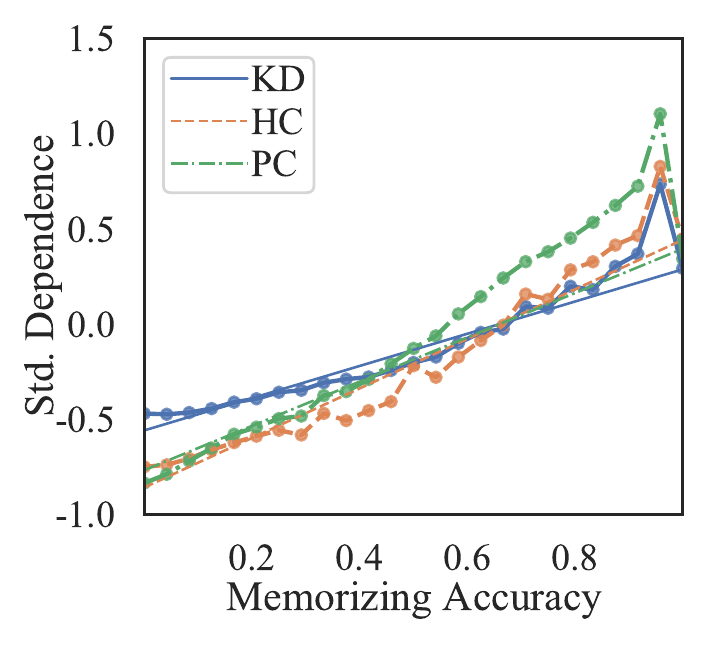}
         \caption{SpanBERT\textit{-large}}
         \label{fig:ate_and_pt_spanbert_large}
     \end{subfigure}\quad
     \begin{subfigure}[b]{0.48\columnwidth}
         \centering
         \includegraphics[width=\textwidth]{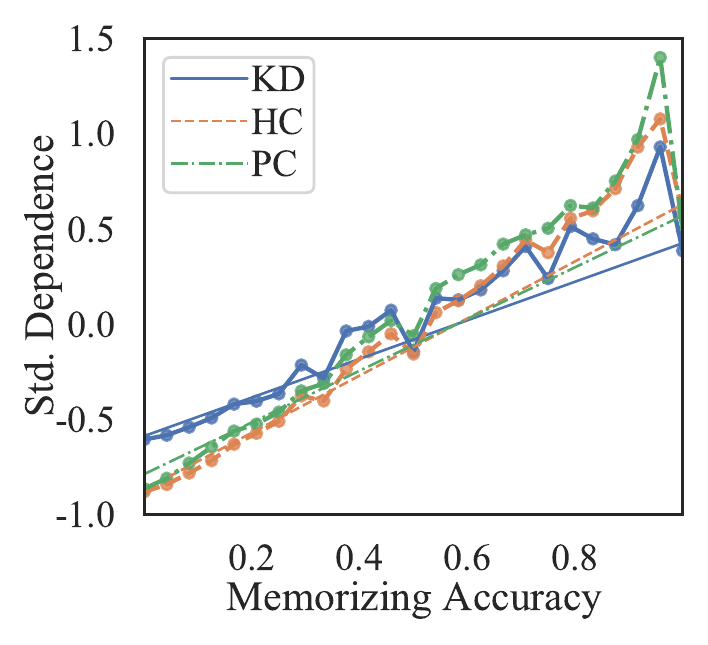}
         \caption{ALBERT\textit{-xxlarge-v2}}
         \label{fig:ate_and_pt_albert_xxlarge_v2}
     \end{subfigure}

\caption{The correlations between the dependence on associations and the mask-filling accuracy on the pre-training samples (Wikipedia sentences).}
\label{fig:ate_and_pt}
\end{figure*}

\subsection{Discussions}
\noindent\textbf{Generality of the Proposed Probing Method}
Generally, the \firstpn offers a way to measure how much the word-level patterns cause the prediction of missing words in Mask Language Model (MLM).  Because words are readable, directly visible, and can be manipulated from the input side directly, the word-level patterns can provide more intuitive interpretations than numeric representation vectors \cite{elazar2021amnesic} or neurons \cite{mediation-investigating}. We use the proposed method to estimate the causal effect of three typical associations in this paper, while this method can be easily adapted to quantify the dependence on any word-level patterns.

\noindent \textbf{Reconsidering ``\textit{PLM as KB}''} 
If we want to use a PLM like a KB, whether the PLM has the same inner workflow as KBs deserves to be considered. The prevalent KBs index knowledge as subject-predicate-object triplets and can infer with triplets \cite{triplet-KB-conceptnet,triplet-KB-freebase,triplet-KB-wikidata}. However, we find out that the knowledge-dependent association, which represents the process of inferring a missing object based on the given subject and predicate, has the lowest dependence in the PLMs. It provides evidence that the PLMs work quite differently with KBs and can not serve stably as KBs for now.

\noindent\textbf{Overfiting and Generalization}
Figure~\ref{fig:ate_and_pt} shows the correlations between the dependence on associations and the mask-filling accuracy on pre-training samples (referred to as memorizing accuracy). The memorizing accuracy increases most as the dependence of the PC association increases, demonstrating that the more PLMs depend on the positionally close words, the better PLMs can recover the pre-training samples. However, there is an opposite trend in probing performance as shown in Figure~\ref{fig:ate_and_query}. The additionally crafted queries used to evaluate the probing performance are mostly unseen in per-training. If we consider these queries as the test set and the pre-training samples as the train set, we can conclude that the dependence on the PC associations makes the PLMs tend to overfit the training data and degrade the generalization on the test set. 

\noindent\textbf{Factual Knowledge Capture in Pre-training}
We want to focus on the pre-training samples that help PLMs to capture factual knowledge. So we reconstruct the pre-training samples that predict some missing factual words (object) based on the factual clues (subject, predicate). We conduct the \firstpn on these samples to investigate how the factual knowledge is captured in pre-training. The mask-filling accuracy on these pre-training samples denotes how well PLMs memorize them in pre-training. We name it as ``train'' in Equation~\ref{eq:correct_metric_train} and ``Memorizing Accuracy'' in Figure~\ref{fig:ate_and_pt}.

\noindent \textbf{Overlap between Associations} The clues for different associations overlap sometimes, e.g., some remaining words may hold the KD and PC associations with the missing words at the same time. The overlaps do not impair the estimations because we use a set of samples to estimate the effect of each association. The samples that hold the same association stay in the same set, and the average causal effect in all these samples is the quantified dependence of this association. The sample sets are quite different for different associations. Table~\ref{tab:dataset_stat_full} shows the corresponding statistics of the overlaps.

\section{Related Works}
\noindent \textbf{Probing Factual Knowledge in PLMs} 
Factual Knowledge Probing in PLMs has attracted much attention recently. LAMA \cite{lama} propose a benchmark that probes the factual knowledge in the PLMs with cloze-style prompts and shows PLMs' ability to capture factual knowledge. This ability can be further explored by tuning the prompts for probing \citet{JiangXAN20,shin2020autoprompt,optim-prompt}.

Motivated by the probing results, some recent works analyze the captured factual knowledge from more perspectives. \citet{lanka} analyze the distribution of predictions and the answer leakage in probing. \citet{EBERT} propose that the PLMs could predict based on some correlation between surface forms rather than infer according to facts. \citet{knowledge-consistency} reveal that the PLMs' outputs are inconsistent as querying the same fact with different prompts. 

This paper proposes a more fine-grained inspection of word-level patterns in the input. In addition to constructing more challenging probing data as input or analyzing the outputs more detailedly, we try to reveal the inner mechanism of PLMs by conducting intervention on the input and then observing the change in the output.

\noindent \textbf{Causal-inspired Interpretations in NLP} 
A causal-inspired approach to explanation is to generate counterfactual examples and then compare the predictions~\cite{feder2021causal}. \citet{feder2021causalm} propose a framework for producing explanation for NLP models using counterfactual representation. 
\citet{mediation-investigating} analyze the effect of neurons (or attention heads) on the gender bias using causal mediation analysis. In this paper, we revisit the word-level post-hot interpretation  \cite{nlp-interpretation-survey,post-hoc-interpretation} from a causal-effect perspective: intervene on some specified words in the input and measure the difference in the output to estimate the causal effect of these words. Furthermore, we evaluate the effectiveness of different causes by calculating correlations between their effects and performance. As far as we know, our work is the first study to probe and evaluate word-level patterns in the factual knowledge capture task.

\section{Conclusion}
In this paper, we try to answer the question of \textit{How Pre-trained Language Models Capture Factual Knowledge} by measuring and evaluating different associations that PLMs use to capture factual knowledge. We present three word-level associations, knowledge-dependent, positionally close, and highly co-occurred in the analysis. The analysis results show that the PLMs rely more on the ineffective positionally close and highly co-occurred associations when capturing factual knowledge, and somewhat ignore the effective knowledge-dependent clues. These findings indicate that we should pay more attention to the knowledge-dependent association to let PLMs capture factual knowledge better.


\bibliography{anthology,custom}
\bibliographystyle{acl_natbib}

\clearpage
\appendix
\section*{Appendix}
\section{Probing Sample Construction for the \FirstPN}
\label{sec:data-detail}
We detail how we construct the probing samples for the \firstpn as follows. We take a subject-predicate-object triplet in Wikidata\footnote{\url{https://www.wikidata.org/}} as a piece of factual knowledge~\cite{lama,lanka,mLAMA,knowledge-consistency}. A subject-predicate-object triplet is aligned with a Wikipedia sentence by matching the subject, predicate, and object with their corresponding spans, respectively. A subject-predicate-object triplet is aligned with a Wikipedia sentence by matching the subject, predicate, and object with their corresponding spans, respectively. The words that correspond to the object are the factual words that are masked and need to be predicted, and we investigate how the different remaining words contribute to the prediction.

The remaining words that have three typical associations with the masked words are considered in the analysis. The rules to identify the remaining words that have the \textit{Knowledge-Dependent}~(\textbf{KD}) association are:
\begin{enumerate}
    \item The $W_o$ and the $W_t$ describe the same subject-predicate-object triplet in KB.
    \item The $W_t$ are the natural language description of the subject and predicate, the $W_o$ are that for the object.
    \item If the subject and predicate corresponding to $W_t$ are given correctly (\ie, $W_t=\hat{w}_t$), the ground-truth value of the object is \textit{unique} in the KB.
\end{enumerate}

The first rule makes the $W_t$ and $W_o$  grounded in the same piece of factual knowledge. The second rule makes predicting the outcome words similar to inferring the object using a KB when giving the subject and predicate. The third rule means if the treatment words are given correctly, there should be one and only one ground-truth value for the object. The third rule is similar to the N-1 relationship \cite{lanka} in KB and lets the ground-truth treatment words can be regarded as a sufficient condition to predict the unique outcome words deterministically.

We use the T-REx\footnote{\url{https://hadyelsahar.github.io/t-rex/}} dataset to provide the initial alignment between KB triplets and the Wikipedia sentences. We use the aliases in KB as keys for fuzzy string match (Levenshtein distance is less than 2, stemming before matching, etc.) to align more subjects, predicates, and objects with spans in the sentence. The sentences that have no aligned triplet are filtered out.

Sometimes, the outcome words in a single sentence relate to multiple triplets that satisfy the rules for KB. E.g., there are two groups of remaining words that can infer the outcome words deterministically based on KB. We select them all as the $W_t$ when probing the KB association and keep the number of the masked words be the same in interventions for the other associations. $D_{\rm KD}$, $D_{\rm PC}$, and $D_{\rm HC}$ denote the sample sets for the Knowledge-Dependent (KD), Positionally Close (PC), and Highly Co-occurred (HC) associations, respectively. 

For the \textit{Highly Co-occurred} association~(\textbf{HC}), the remaining words that are top-$k$ in PMI with the ground-truth outcome words $\hat{w}_o$ are selected as $W_t$. The $k$ is the number of words with the KD associations for the same sentence. The PMI between words is calculated by all the Wikipedia sentences. If the $\hat{w}_o$ consists of multiple words, occurring with all the words in $\hat{w}_o$ altogether are taken as co-occurring. The order of the words in $\hat{w}_o$ are ignored for efficiency. Table~\ref{tab:dataset_stat_full} shows more details about the probing samples.

\begin{table*}[htbp]
\small
\centering
\begin{tabular}{lc}
\toprule
\textbf{Probing Samples in \FirstPN} &\\
\midrule
\# Average treatment words     & 4.0023  \\
\# Average outcome words       & 1.8588  \\
\# Average words               & 23.1031 \\
\midrule\midrule


\textbf{Word-level Overlap Between Associations} \\ 
\midrule
$D_{\rm KD} \cap D_{\rm HC}$                  & 44.25\% (8,455,641) \\
$D_{\rm KD} \cap D_{\rm PC}$                  & 20.83\% (3,981,305) \\
$D_{\rm PC} \cap D_{\rm HC}$                  & 18.75\% (3,582,576)\\
$D_{\rm KD} \cap D_{\rm PC} \cap D_{\rm HC}$  & 8.3349\% (1,592,560) \\
\midrule\midrule
\textbf{Sample-level Overlap Between Associations} \\ 
\midrule
$D_{\rm KD} \cap D_{\rm HC}$                  & 3.12\% (149,479) \\
$D_{\rm KD} \cap D_{\rm PC}$                  & 0.06\% (2,995) \\
$D_{\rm PC} \cap D_{\rm HC}$                  & 0.12\% (5,777)\\
$D_{\rm KD} \cap D_{\rm PC} \cap D_{\rm HC}$  & 0.0001\% (547) \\
\midrule\midrule
\textbf{Template-based Queries in \SecondPN} \\
\midrule
    \# Average treatment words     & 1.9484  \\
    \# Average outcome words       & 1.5844  \\
    \# Average word per sample     & 6.9617  \\

\bottomrule
\end{tabular}
\caption{Statistic of the probing data in the \firstpn.}
\label{tab:dataset_stat_full}
\end{table*}

\section{More Probing Results}


The Pearson correlation coefficients between the dependence on associations (raw value without standardization) and the performance are shown in Table~\ref{tab:ate_and_query_full}. The three metrics, accuracy (defined in Equation~\ref{eq:correct_metric}), consistency (defined in Equation~\ref{eq:consist_metric}),  and the overall performance metric (defined in Equation~\ref{eq:joint_metric}), are reported respectively. The correlation coefficients between the dependence and the performance are consistent with the slopes of the regression lines in Figure~\ref{fig:ate_and_query}. Table~\ref{tab:acc_desc} shows the accuracy decreasing results after masking the treatments words when generating the missing words in Wikipedia sentences.

\begin{table*}[ht]
    \small
    \centering
    \begin{tabular}{lccccc}
    \toprule
     \multirow{2}{*}{\textbf{Model}} & \multicolumn{5}{c}{\textbf{Input Context} (Accuracy, \%)} \\ 
     \cmidrule{2-6} 
                                    & Complete &       w/o \textbf{KD} &         w/o \textbf{PC} &        w/o \textbf{HC} & w/o \textbf{R}    \\
    \midrule
    BERT\textit{-base-cased}        &    36.23 &        22.05 (-14.17) &         \ 2.67 (-33.56) &         19.68 (-16.54) &   26.71 (\ -9.52) \\
    BERT\textit{-large-cased}       &    36.92 &        22.11 (-14.81) &         \ 2.70 (-34.22) &         19.04 (-17.88) &   27.11 (\ -9.81) \\
    BERT\textit{-large-cased-wwm}   &    50.30 &        35.22 (-15.08) &         11.50 (-38.80)  &         26.09 (-24.21) &   39.04 (-11.25)  \\
    SpanBERT\textit{-large}         &    52.23 &        36.87 (-15.35) &         16.66 (-35.57)  &         26.78 (-25.44) &   39.87 (-12.35)  \\
    RoBERTa\textit{-base}           &    35.11 &        23.07 (-12.03) &         \ 4.81 (-30.30) &         16.38 (-18.72) &   25.70 (\ -9.41) \\
    RoBERTa\textit{-large}          &    42.76 &        28.26 (-14.50) &         \ 8.98 (-33.78) &         21.21 (-21.54) &   32.67 (-10.09)  \\
    
    \hline
    
    BERT\textit{-base-uncased}      &    36.90 &        23.53 (-13.37) &         \ 3.33 (-33.58) &         20.78 (-16.12) &   28.53 (\ -8.37) \\
    BERT\textit{-large-uncased}     &    38.62 &        24.58 (-14.04) &         \ 2.88 (-35.74) &         21.61 (-17.00) &   29.83 (\ -8.79) \\
    BERT\textit{-large-uncased-wwm} &    50.35 &        35.03 (-15.31) &         11.92 (-38.43)  &         26.16 (-24.18) &   39.03 (-11.32)  \\
    ALBERT\textit{-xxlarge-v2}      &    47.58 &        23.02 (-24.56) &         11.58 (-36.00)  &         15.25 (-32.34) &   24.42 (-23.16)  \\
    ALBERT\textit{-xlarge-v2}       &    40.87 &        16.13 (-24.74) &         \ 8.15 (-32.72) &         \ 9.57 (-31.30)&   15.11 (-25.76)  \\
    ALBERT\textit{-large-v2}        &    39.02 &        \ 8.03 (-30.99)&         \ 3.82 (-35.20) &         \ 4.52 (-34.50)&   \ 9.00 (-30.02) \\
    ALBERT\textit{-base-v2}         &    32.76 &        \ 3.63 (-29.13)&         \ 1.74 (-31.03) &         \ 1.76 (-31.00)&   \ 3.63 (-29.13) \\
    ALBERT\textit{-xxlarge-v1}      &    47.50 &        23.68 (-23.83) &         12.20 (-35.30)  &         15.79 (-31.71) &   25.27 (-22.24)  \\
    ALBERT\textit{-xlarge-v1}       &    48.19 &        21.15 (-27.04) &         11.81 (-36.38)  &         12.82 (-35.38) &   22.57 (-25.62)  \\
    ALBERT\textit{-large-v1}        &    43.64 &        14.14 (-29.50) &         \ 7.52 (-36.12) &         \ 7.93 (-35.71)&   14.80 (-28.84)  \\
    ALBERT\textit{-base-v1}         &    40.95 &        27.39 (-13.56) &         13.13 (-27.83)  &         19.30 (-21.65) &   30.05 (-10.90)  \\
    
    \bottomrule
    \end{tabular}
    \caption{The accuracy of the predictions when different treatment words are missing.}
    \label{tab:acc_desc}
\end{table*}

\begin{table*}[htbp]
    \small
    \centering
    \begin{tabular}{lcccc}
    \toprule
    \multirow{2}{*}{\textbf{Model}} & \multicolumn{4}{c}{\textbf{ATE of Association} ($k=100$/$k=1$)}                                 \\ \cmidrule{2-5} 
                                    &  \textbf{KD}      &        \textbf{PC} &        \textbf{HC} &  \textbf{R}    \\
    \midrule
    
    BERT\textit{-base-cased}        &     0.1585/0.1416 &      0.4085/0.3354 &      0.1779/0.1654 &  0.1081/0.0950 \\
    BERT\textit{-large-cased}       &     0.1603/0.1480 &      0.4113/0.3420 &      0.1791/0.1788 &  0.0996/0.0979 \\
    BERT\textit{-large-cased-wwm}   &     0.1384/0.1506 &      0.4477/0.3879 &      0.2305/0.2421 &  0.1072/0.1124 \\
    SpanBERT\textit{-large}         &     0.1351/0.1533 &      0.3679/0.3555 &      0.2383/0.2544 &  0.1157/0.1233 \\
    RoBERTa\textit{-large}          &     0.1360/0.1438 &      0.3962/0.3366 &      0.2162/0.2154 &  0.0985/0.0997 \\
    RoBERTa\textit{-base}           &     0.1352/0.1192 &      0.3926/0.3018 &      0.2093/0.1872 &  0.1053/0.0929 \\
    \hline
    BERT\textit{-base-uncased}      &     0.1439/0.1337 &      0.4112/0.3357 &      0.1643/0.1612 &  0.0901/0.0837 \\
    BERT\textit{-large-uncased}     &     0.1522/0.1403 &      0.4401/0.3573 &      0.1713/0.1700 &  0.0946/0.0879 \\
    BERT\textit{-large-uncased-wwm} &     0.1410/0.1531 &      0.4350/0.3842 &      0.2290/0.2418 &  0.1089/0.1131 \\
    
    ALBERT\textit{-base-v2}         &     0.3987/0.2911 &      0.4269/0.3100 &      0.4269/0.3100 &  0.4021/0.2911 \\
    ALBERT\textit{-large-v2}        &     0.4075/0.3098 &      0.4716/0.3519 &      0.4566/0.3450 &  0.3958/0.3001 \\
    ALBERT\textit{-xlarge-v2}       &     0.3279/0.2474 &      0.4336/0.3272 &      0.4170/0.3130 &  0.3468/0.2576 \\
    ALBERT\textit{-xxlarge-v2}      &     0.2852/0.2457 &      0.4338/0.3601 &      0.3801/0.3234 &  0.2704/0.2317 \\
    
    ALBERT\textit{-base-v1}         &     0.1429/0.1355 &      0.3235/0.2782 &      0.2287/0.2165 &  0.1167/0.1089 \\
    ALBERT\textit{-large-v1}        &     0.3638/0.2951 &      0.4569/0.3612 &      0.4488/0.3571 &  0.3621/0.2884 \\
    ALBERT\textit{-xlarge-v1}       &     0.3223/0.2705 &      0.4425/0.3639 &      0.4266/0.3538 &  0.3116/0.2563 \\
    ALBERT\textit{-xxlarge-v1}      &     0.2761/0.2384 &      0.4230/0.3531 &      0.3708/0.3171 &  0.2590/0.2224 \\

    \bottomrule
    \end{tabular}
    \caption{The ATE of associations in more PLMs.}
    \label{tab:ate_full}
\end{table*}

\begin{table*}[htbp]
\centering
\tiny
\begin{tabular}{lcccc}
\toprule
\multirow{2}{*}{\textbf{Model}}    & \multicolumn{4}{c}{\textbf{Associations} (joint/accuracy/consistency)}                                      \\ \cmidrule{2-5}
                                   & \textbf{KD}    &  \textbf{PC} &  \textbf{HC} &  \textbf{R} \\
\midrule

BERT\textit{-base-cased}           &  0.1523/0.2222/0.0768 &  -0.2156/-0.1839/-0.1597 &   \ 0.0011/\ 0.0180/-0.0137  &  -0.0823/-0.0774/-0.0621 \\
BERT\textit{-large-cased}          &  0.1398/0.1879/0.0788 &  -0.1638/-0.1120/-0.1295 &   -0.0017/\ 0.0095/-0.0141   &  -0.0810/-0.0741/-0.0638 \\
BERT\textit{-large-cased-wwm}      &  0.2492/0.2795/0.1627 &  -0.1904/-0.1783/-0.1290 &   \ 0.0800/\ 0.0851/\ 0.0422 &  -0.0487/-0.0455/-0.0417 \\
SpanBERT\textit{-large}            &  0.2463/0.2784/0.1382 &  -0.1068/-0.0781/-0.1187 &   \ 0.1017/\ 0.1175/\ 0.0254 &  -0.0384/-0.0307/-0.0391 \\
RoBERTa\textit{-base}              &  0.2432/0.3062/0.1223 &  -0.0414/-0.0440/-0.0366 &   \ 0.0966/\ 0.1252/\ 0.0297 &  -0.0201/-0.0054/-0.0423 \\
RoBERTa\textit{-large}             &  0.2212/0.2666/0.1141 &  -0.1131/-0.1455/-0.0642 &   \ 0.0749/\ 0.0911/\ 0.0156 &  -0.0311/-0.0441/-0.0236 \\\midrule
BERT\textit{-base-uncased}         &  0.1635/0.2233/0.0954 &  -0.1454/-0.1269/-0.1182 &   \ 0.0130/\ 0.0410/-0.0114  &  -0.0659/-0.0630/-0.0596 \\
BERT\textit{-large-uncased}        &  0.1507/0.2022/0.0749 &  -0.2056/-0.1900/-0.1084 &   \ 0.0247/\ 0.0355/\ 0.0085 &  -0.0671/-0.0667/-0.0454 \\
BERT\textit{-large-uncased-wwm}    &  0.2526/0.2776/0.1593 &  -0.1772/-0.1589/-0.1346 &   \ 0.0866/\ 0.0886/\ 0.0344 &  -0.0462/-0.0419/-0.0401 \\
ALBERT\textit{-base-v2}            &  0.0453/0.0530/0.0347 &  -0.1054/-0.1333/-0.0417 &   \ 0.0071/-0.0005/\ 0.0371  &  -0.0886/-0.1186/-0.0117 \\
ALBERT\textit{-large-v2}           &  0.0809/0.0988/0.0370 &  -0.1130/-0.1457/-0.0826 &   \ 0.0201/\ 0.0158/\ 0.0093 &  -0.0822/-0.1061/-0.0707 \\
ALBERT\textit{-xlarge-v2}          &  0.1515/0.2161/0.1064 &  -0.1184/-0.1152/-0.0759 &   \ 0.0278/\ 0.0524/\ 0.0154 &  -0.0997/-0.0978/-0.0627 \\
ALBERT\textit{-xxlarge-v2}         &  0.1685/0.1954/0.1347 &  -0.1549/-0.1552/-0.1039 &   \ 0.0445/\ 0.0492/\ 0.0496 &  -0.0783/-0.0759/-0.0559 \\
ALBERT\textit{-base-v1}            &  0.3034/0.3563/0.1764 &  -0.0650/-0.0504/-0.1010 &   \ 0.1564/\ 0.1724/\ 0.0497 &  \ 0.0111/\ 0.0175/-0.0202 \\
ALBERT\textit{-large-v1}           &  0.1466/0.1741/0.1145 &  -0.0384/-0.0639/-0.0037 &   \ 0.0598/\ 0.0530/\ 0.0606 &  -0.0186/-0.0408/\ 0.0026 \\
ALBERT\textit{-xlarge-v1}          &  0.1593/0.1816/0.1486 &  -0.0514/-0.0629/-0.0006 &   \ 0.0498/\ 0.0338/\ 0.1172 &  -0.0081/-0.0288/\ 0.0431 \\
ALBERT\textit{-xxlarge-v1}         &  0.1926/0.2207/0.1462 &  -0.1314/-0.1346/-0.0929 &   \ 0.0629/\ 0.0647/\ 0.0526 &  -0.0563/-0.0535/-0.0475 \\


\bottomrule

\end{tabular}

\caption{The Pearson correlation coefficients between the ATEs and the factual knowledge capture metrics.}
\label{tab:ate_and_query_full}
\end{table*}

\end{document}